# Crash Narrative-Guided Countermeasure Recommendation Using Large Language Models: A Retrieval-Augmented Generation Framework for Intersection Safety

**Abu Saif Md Nasim Uddin, PhD Student**
Department of Civil, Environmental & Construction Engineering,
University of Central Florida, Orlando, FL 32816, United States
E-mail: ab038808@ucf.edu

**Mohamed Abdel-Aty, Professor**
Department of Civil, Environmental & Construction Engineering,
University of Central Florida, Orlando, FL 32816, United States
Email: M.aty@ucf.edu

**Zubayer Islam, Assistant Professor**
Department of Civil, Environmental & Construction Engineering,
University of Central Florida, Orlando, FL 32816, United States
Email: Zubayer.Islam@ucf.edu

**Parvez Anowar, PhD Student**
Department of Civil, Environmental & Construction Engineering,
University of Central Florida, Orlando, FL 32816, United States
Email: pa545735@ucf.edu

**Chenzhu Wang, Postdoctoral Researcher**
Department of Civil, Environmental & Construction Engineering,
University of Central Florida, Orlando, FL 32816, United States
Email: chenzhu.wang@ucf.edu

**Abstract**

Improving safety at intersections requires identifying crash mechanisms and recommending appropriate countermeasures. However, this process traditionally relies on expert judgment, making it labor-intensive, difficult to scale, and dependent on the availability of experienced traffic safety engineers. Although crash narratives contain rich description of crash mechanisms, this unstructured information remains largely underutilized in safety analyses. This study presents a crash narrative-guided retrieval-augmented generation (RAG) framework that translates narrative-derived crash mechanisms into site-specific countermeasure recommendations. Key mechanism attributes including traffic control, signal indication, driver fault, vehicle movement, and travel direction were extracted from crash narratives and linked to evidence-based treatments from the FHWA Proven Safety Countermeasures and the CMF Clearinghouse. The framework integrates embedding-based retrieval of historically similar intersections, association-rule mining, statistical guidance on the expected number of relevant countermeasures, and an engineering reasoning guidance that directs LLM through a domain-consistent decision process before selecting countermeasures. Evaluated on 312 fatal and serious-injury crashes across 115 intersections in Lake and Sumter Counties, Florida, using five-fold cross-validation, the framework achieved a precision of 0.82, recall of 0.85, and F1-score of 0.82, while recommending an average of 3.91 countermeasures per location with 3.14 matching, closely matching the actual average (3.86). Overall, the proposed framework demonstrates the potential of retrieval-augmented LLMs as an interpretable and scalable decision-support tool for transportation agencies for translating crash narratives into countermeasure recommendations.

## 1 Introduction

Intersections remain one of the most critical components of the roadway network from a traffic safety perspective in the US. According to the Federal Highway Administration (FHWA 2024) statistics derived from NHTSA's Fatality Analysis Reporting System (FARS) indicate that, during 2018–2022, approximately 28% of all traffic fatalities in the United States occurred at or were related to intersections. Among these fatalities, nearly 66% occurred at unsignalized intersections, while approximately 34% occurred at signalized intersections. Although intersections are small proportion of the roadway system, they are inherently complex due to the diverse behaviors and interactions of road users, including turning vehicles, through traffic, pedestrians, and bicyclists, which create a high potential for severe conflicts (Shirazi and Morris 2017). Therefore, improving intersection safety requires not only identifying high-risk locations but also recommending appropriate countermeasures, engineering interventions proven to reduce crash frequency and severity in a timely, evidence-based, and site-specific manner.

Traditionally, countermeasure selection has relied on the expertise of traffic safety engineers who analyze crash data, consult resources such as the Highway Safety Manual (AASHTO 2010) and the FHWA Crash Modification Factors Clearinghouse (CMF Clearinghouse 2025), and apply professional judgment to match observed crash patterns with appropriate treatments. While this process draws on deep domain knowledge, it is resource-intensive, difficult to scale across large road networks, and inherently subject to the availability and experience of qualified personnel. As transportation agencies face a growing need to address safety systematically, particularly under the Safe System Approach for eliminating roadway fatalities, there is increasing interest in automated, data-driven methods that can assist engineers in prioritizing and selecting effective countermeasures at scale. The effectiveness of such approaches, however, depends on their ability to leverage all available sources of crash information. Despite the growing availability of detailed crash records, valuable information contained in crash narratives and diagrams is often underutilized in conventional safety analyses, which primarily depend on aggregated crash statistics and structured variables (Wu et al. 2026; Fitzpatrick et al. 2017).

A crash narrative is the responding officer's written account describing the sequence of events before, during, and after a crash (Fitzpatrick et al. 2017). In crash databases, crash events are typically represented through coded fields such as crash type, injury severity, roadway condition, intersection type, and behavioral indicators such as distracted driving, DUI, or speeding. Although these variables support statistical modeling and network-level safety screening, they simplify the crash process into predefined categories and often fail to capture the sequence of vehicle movements, driver actions, traffic control devices, and immediate pre-crash circumstances. Crash narratives therefore provide an important unstructured source for examining crash mechanisms beyond what is captured in tabulated records. **Figure 1** presents an example of how coded crash data, crash narratives, and crash diagrams provide different levels of information. The coded record identifies the event as an angle crash at a signalized four-way intersection during daylight conditions, resulting in an incapacitating injury. However, the narrative and diagram provide additional details: the northbound through vehicle was proceeding on a green signal, while the southbound left-turning vehicle was operating under a flashing yellow arrow and failed to yield. Therefore, the crash mechanism is not simply an angle crash at a signalized intersection, but a left-turn failure-to-yield crash under permissive left-turn phasing. If such patterns recur at a location, they may indicate the need for targeted countermeasures such as protected-only left-turn phasing or other treatments addressing left-turn conflicts.

However, extracting useful information from these unstructured textual descriptions traditionally requires manual examination of crash reports and associated diagrams, making the process labor-intensive. Recent advances in natural language processing (NLP) have created new opportunities to automatically analyze unstructured textual data and uncover information that was previously difficult to utilize in large-scale safety analyses. However, traditional machine learning and NLP approaches are often limited in their ability to capture the semantic and contextual meaning of text, restricting their understanding of complex crash descriptions (Mumtarin et al. 2023).

More recently, large language models (LLMs), built upon transformer architectures, have demonstrated remarkable capabilities in understanding context, reasoning over complex textual descriptions, making them particularly promising for extracting crash-related information and supporting safety decision-making (Karim et al. 2025; Abdelrahman et al. 2025). Nevertheless, because LLMs rely heavily on learned parametric knowledge, they may generate outputs that are inaccurate, unsupported, or unfaithful to the provided source information (Ji et al. 2023). Retrieval-Augmented Generation (RAG) has emerged as a powerful framework for grounding LLM outputs in domain-specific evidence (Lewis et al. 2020; Gao et al. 2024). Rather than relying solely on a model's parametric knowledge, RAG systems retrieve relevant information from an external knowledge base and use it to guide generation, thereby improving factual accuracy, reducing hallucinations, and enabling responses to be grounded in retrieved evidence (Lewis et al. 2020; Asai et al. 2024). In the context of intersection safety, RAG provides a natural fit, where a target location's crash narrative can be used to retrieve historically similar intersections and leverage their associated countermeasures to support recommendations for the new site.

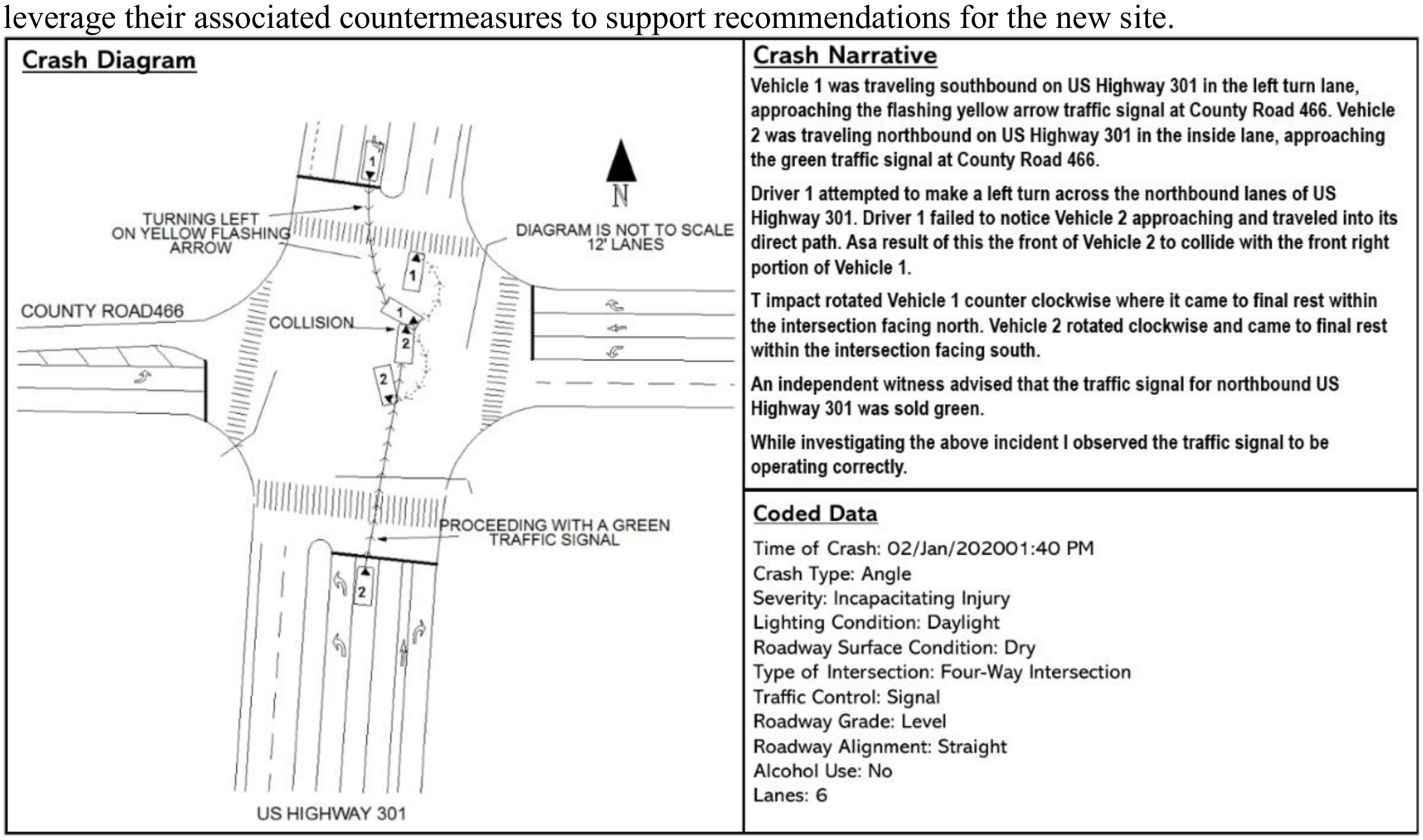

**Figure 1. Example of complementary information provided by coded data, crash narratives, and crash diagrams**

Despite the availability of rich narrative data and recent methodological advances, the use of crash narratives for safety countermeasure recommendation remains limited. Existing narrative-based safety studies have primarily focused on contributing-factor identification, or scenario extraction (Hossain et al. 2024; Li et al. 2024), while comparatively little attention has been given to translating narrative-derived crash mechanisms into actionable engineering countermeasures. Motivated by this gap, this study develops a crash narrative-guided retrieval-augmented large language model framework for intersection countermeasure recommendation. The study contributes by:

- Extracts key information from unstructured crash narratives, including traffic control, signal indication, driver fault, vehicle movements, and travel directions of involved vehicles, to provide a comprehensive representation of crash mechanisms.
- Develops a location-level crash-countermeasure knowledge base for 115 intersections in Lake and Sumter Counties, Florida, by linking narrative-derived crash mechanisms with evidence-based treatments from the FHWA Proven Safety Countermeasures and the Crash Modification Factors (CMF) Clearinghouse.

- Develops and evaluates a crash narrative-guided retrieval-augmented generation framework that leverages historically similar intersections to generate site-specific countermeasure recommendations.

The remainder of this paper is organized as follows. Section 2 reviews related work on crash narrative analysis, countermeasure recommendation, and the application of LLMs in transportation safety. Section 3 describes the dataset and data preparation methodology. Section 4 presents the proposed RAG framework in detail. Section 5 reports evaluation results and discusses performance across crash types and intersection configurations. Section 6 concludes with implications for practice and directions for future research.

## 2 Literature Review

### 2.1 Crash Narratives in Traffic Safety

Early applications of crash narratives were largely descriptive and relied on manual review or keyword-based approaches. Sorock et al. (1996) used keyword frequency and predefined crash categories to examine work-zone crash narratives and identify recurring hazards. Similarly, Pollack et al. (2013) reviewed military vehicle crash narratives to identify crash determinants and assess the potential role of driving assistance technologies. Other studies used narratives to examine specific crash contexts, including young driver crashes, motorcyclist crashes, and distraction-related crashes (McKnight and McKnight 2003; Graves et al. 2015; Dube et al. 2016). As the availability of crash narratives increased, researchers gradually moved from manual examination toward automated text mining and machine-learning approaches. These methods were used to uncover contributing factors, crash theme extraction and classification, predict secondary crashes (Nayak et al. 2010; Zhang et al. 2020; Kwayu et al. 2021).

More recent work has moved toward more structured information extraction from crash narratives. Natural language processing techniques such as keyword extraction and named entity recognition have been used to extract variables such as vehicle type, travel direction, driver violation, hazardous action, and initial collision position (Sayed et al. 2021; Yan et al. 2023). Kwayu et al. (2020) used crash narratives to detect hazardous driver actions at signalized intersections, while Li et al. (2024) used dependency parsing-based information extraction to support crash scene reconstruction. Other recent studies have used narrative-derived information for crash diagram generation, multimodal crash mechanism analysis, injury severity prediction, or contributing-factor (Lu et al. 2026; Wu et al. 2026; Shao et al. 2024; Mumtarin et al. 2023).

Overall, the literature shows the growing use of crash narratives in safety analysis, but no previous work has leveraged narrative-derived crash mechanisms to support site-specific engineering decisions, particularly the recommendation of intersection countermeasures. This gap is important because effective countermeasure selection requires translating crash mechanisms into treatments that can prevent similar crashes in the future.

### 2.2 Decision Support in Countermeasure Recommendation

Although road safety research has devoted considerable attention to evaluating countermeasure effectiveness, comparatively fewer studies have addressed the process of generating or recommending countermeasures themselves. Existing efforts have employed multicriteria decision-support systems for prioritizing safety interventions (Fancello et al. 2015), knowledge-based frameworks linking crash causes and countermeasures (Martensen et al. 2019), high-level evidence-based policy frameworks connecting crash causation to countermeasure development (Talbot et al. 2024). More recently, machine-learning methods have been investigated for automated countermeasure recommendation (Park et al. 2025). Nevertheless, several researchers have emphasized that effective prevention requires moving beyond simply identifying crash causes and instead focusing on interventions that can prevent similar crashes in the future (Shinar 2019; Shinar and Hauer 2024). Overall, there remains substantial scope for developing more effective decision-support tools for countermeasure recommendation.

### 2.3 Large Language Models in Transportation

Recent advances in large language models (LLMs) have led to increasing applications of generative artificial intelligence in transportation engineering. Studies have explored the use of LLMs for travel behavior analysis (Gong et al. 2024; Mo et al. 2026), autonomous driving (Cui et al. 2024; Fu et al. 2024), aviation (Liu 2024; Wang et al. 2024), traffic management and control (Movahedi and Choi 2025; Mahmud et al. 2025), and transportation planning and mobility analysis (Zhang et al. 2026). In traffic safety, LLMs have been increasingly employed for crash analysis and safety diagnostics. (Wang et al. 2023) introduced AccidentGPT, a multimodal framework integrating V2X perception data with GPT-4V for accident analysis and automated report generation, while Fan et al. (2024) developed CrashLLM, which treats crash analysis as a language reasoning task and enables crash outcome prediction and causal what-if analyses. Recent studies have also combined LLMs with video-based detection systems to analyze crash and near-miss events and support proactive safety interventions (Jaradat et al. 2025, 2024). For improved domain adaptation, researchers have explored fine-tuning LLMs for transportation-specific tasks such as trajectory prediction, time-series forecasting, and the development of specialized models (Chang et al. 2025; Lan et al. 2025; Ren et al. 2024). However, because fine-tuning requires substantial labeled data and computational resources, prompt engineering has emerged as an effective alternative for task-specific adaptation. More recently, Retrieval-Augmented Generation (RAG) has enabled LLMs to incorporate domain-specific knowledge from external sources, making them particularly suitable for safety-critical transportation applications (Karim et al. 2025).

Overall, the literature highlights the remarkable capabilities of LLMs in performing complex transportation tasks. In traffic safety, these capabilities may extend beyond crash analysis to supporting engineering decisions, making LLMs a promising foundation for intelligent decision-support systems for safety countermeasure recommendation.

## 3 Methodology

### 3.1 Data Description

The study utilized crash data for 2020–2024 from Signal Four Analytics (S4A) for Lake and Sumter Counties, Florida. A total of 312 fatal and incapacitating injury (KA) crashes occurring at 115 intersection locations were included in the analysis. Among the study locations, 58 were stop-controlled, 52 were signalized, and 5 were uncontrolled. Each crash record consisted of both structured variables and unstructured information in the form of officer-written narratives and, when available, crash diagrams.

### 3.2 Narrative Processing and Countermeasure Knowledge Base Development

Crash narratives and diagrams were interpreted with the assistance of GPT-5.5. The model was employed to facilitate the extraction of mechanism-related attributes from the crash reports, while the resulting information was reviewed and standardized to ensure consistency and accuracy. Among the extracted information, vehicle travel direction was particularly important because it helped identify the interaction and conflict patterns between road users. **Figure** illustrates several examples of left-turn crash patterns classified based on vehicle maneuvers. (Wang and Abdel-Aty 2008b) reported that left-turn and angle crashes account for the highest proportion of intersection crashes and identified nine distinct crash patterns associated with different crash mechanisms. For example, Patterns (2) and (4) in **Figure** were found to be among the most frequent and severe. Because each pattern reflects a different crash mechanism, the appropriate countermeasure may vary depending on the vehicle interactions, intersection configuration, and traffic control (Wang and Abdel-Aty 2008a).

Signal indications were also extracted from crash narratives, when available. For signalized intersections, the operational status of the traffic signal at the time of the crash provided important context for identifying crash mechanisms and assigning fault. Similarly, for stop-controlled intersections, information regarding stop sign compliance and right-of-way violations helped identify the contributing circumstances of the crash. For example, crashes involving red-light running, permissive left-turn conflicts under flashing yellow arrows, and failures to yield while turning may result in similar collision types but arise from fundamentally different safety issues. Consequently, signal indications helped distinguish between these conditions and identify context-appropriate countermeasures. Further, driver violations and contributing actions extracted from the crash narratives provided insight into the behavioral factors associated with the crash, such as failure to yield, red-light running, stop-sign violations, distracted driving, following too closely, and improper maneuvers. These actions and violations often reflect different underlying safety issues and may warrant different countermeasures. Additionally, the crash narratives were used to cross-check the recorded crash type, intersection type, traffic control, and lighting condition for potential inconsistencies. During the review process, it was found that a substantial number of crashes were misclassified or recorded as "Other". Information contained in the narratives was therefore used to verify and, when necessary, correct these attributes to improve the consistency and accuracy of the dataset.

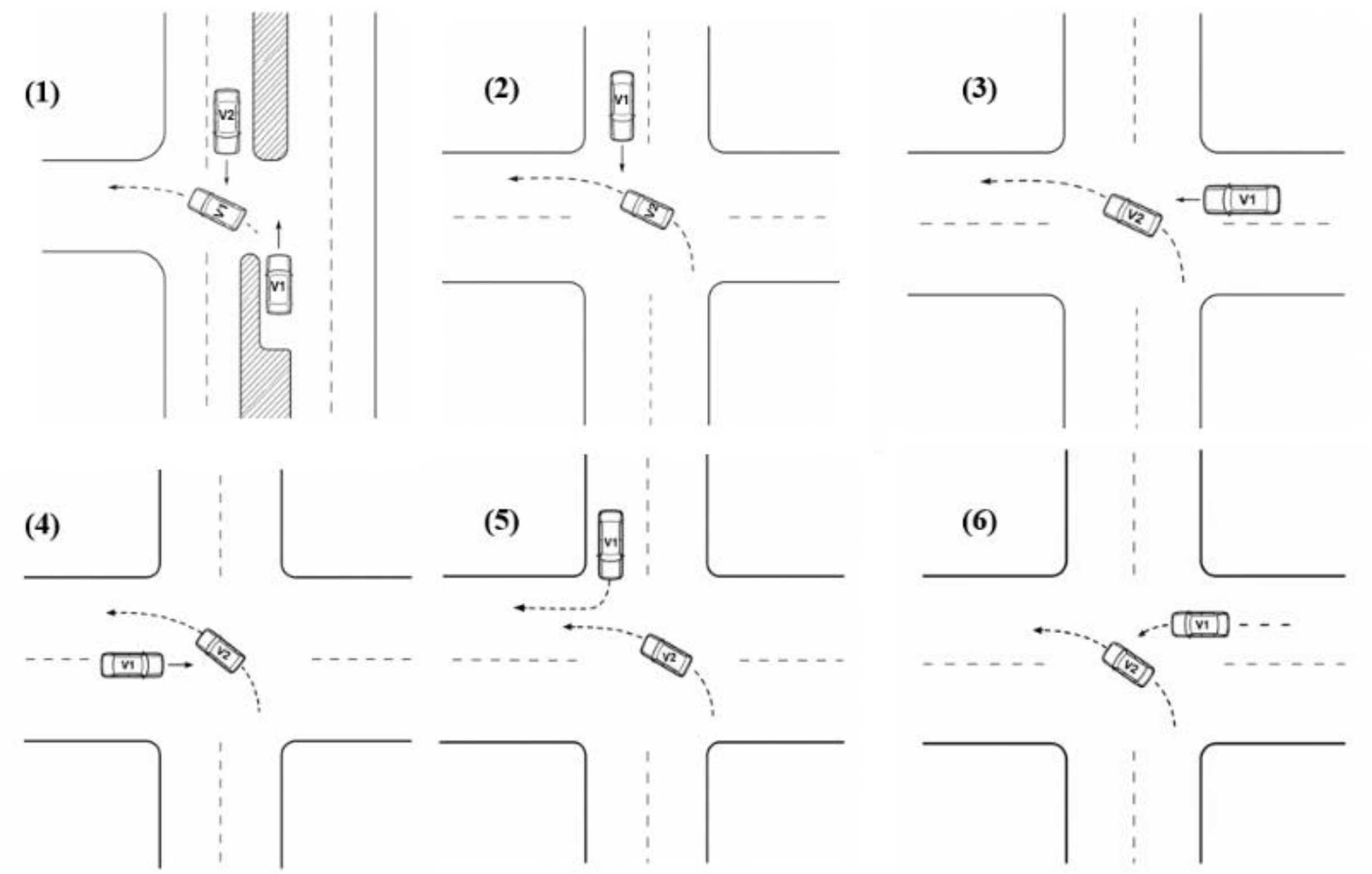

**Figure 2. Example of common left-turn crash patterns based on vehicle maneuvers**

The extracted information was subsequently used to identify recurring crash patterns and develop location-specific safety interventions. **Figure** illustrates the overall process used to translate crash narratives into countermeasure recommendation. For each intersection, crash reports and narratives were reviewed to identify dominant crash types, contributing factors, and recurring conflict patterns. Crash-specific mechanisms were then synthesized at the location level to determine the primary safety issues affecting the intersection. Finally, candidate treatments were selected from the CMF Clearinghouse and FHWA Proven Safety Countermeasures based on the identified crash patterns and contributing factors. For example, at the intersection of US 301 and CR 466 in Sumter County, a review of crashes occurring between 2020 and 2024 identified five crashes in which a southbound left-turning vehicle failed to yield under a flashing yellow arrow and entered the path of a northbound through vehicle proceeding on a green signal. In addition, three incapacitating injury crashes were associated with red-light running violations. To reduce the likelihood of similar crashes, countermeasures such as converting the left-turn movement to protected-only phasing (CMF = 0.239 for left-turn crashes), signal timing adjustments including yellow interval optimization (CMF = 0.67 for red-light-running crashes), and the installation of red-light enforcement cameras (CMF = 0.82 for left-turn crashes) were recommended for the intersection.

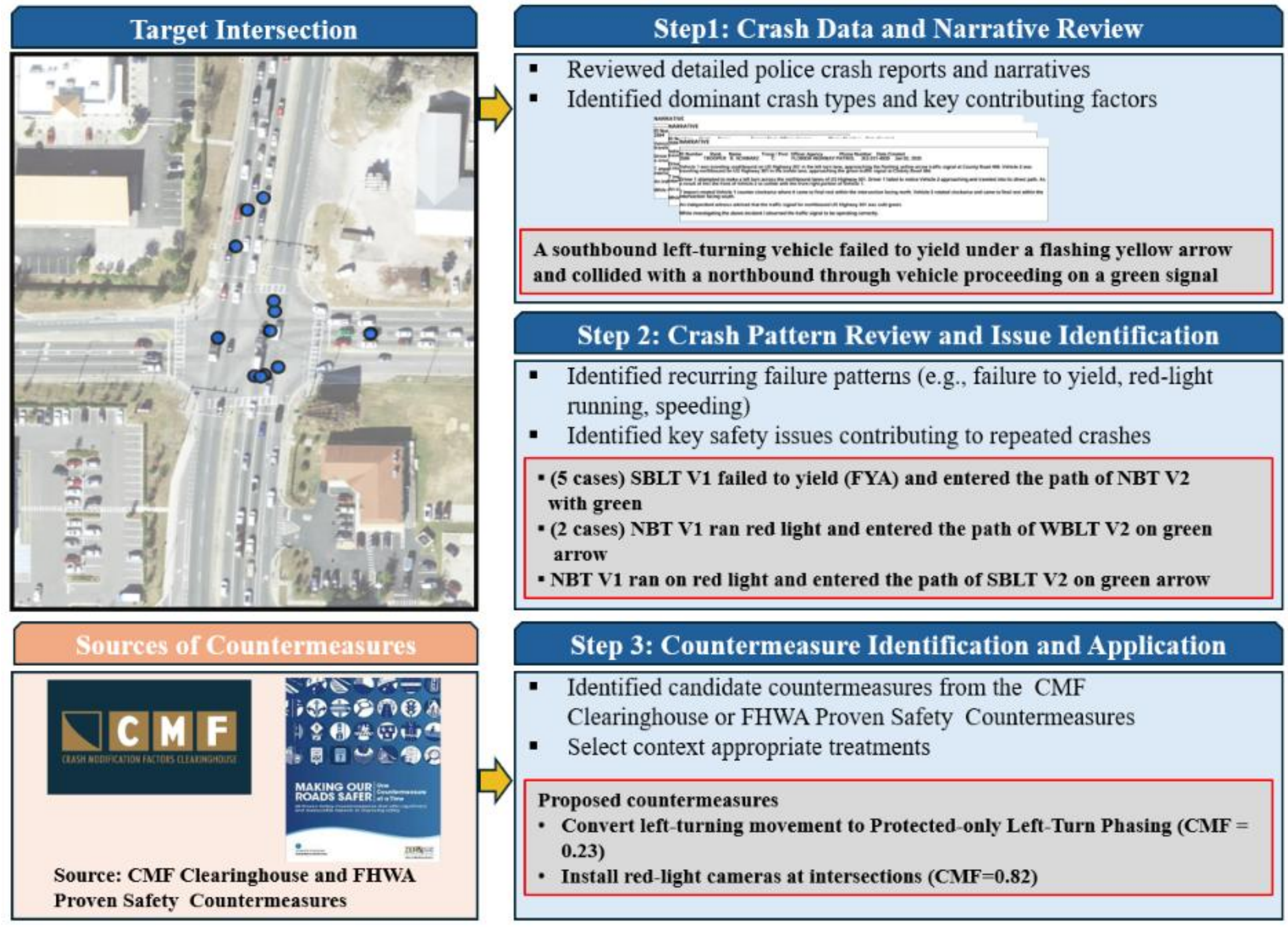

**Figure 3. Process for developing countermeasure recommendations at an intersection**

### 3.3 Training Data Preparation

The narrative-derived crash database and corresponding countermeasure database developed in the previous step were subsequently used to prepare the training data for the proposed countermeasure recommendation framework. **Figure** illustrates the overall training data preparation process. The 115 intersection locations were divided into 92 training locations and 23 test locations. The training locations were used to construct three complementary knowledge components: an embedding-based retrieval database, a master countermeasure list, and association-rule-based countermeasure guidance.

Sentence embeddings were generated using the all-MiniLM-L6-v2 model from the Sentence Transformers framework, which combines the Sentence-BERT architecture for semantic similarity learning with the lightweight MiniLM transformer for efficient representation learning (Reimers and Gurevych 2019; Wang et al. 2020). For each training location, crash narratives and key structured descriptors were concatenated into a single textual representation, denoted as $s_i$. The resulting representation was encoded into a dense embedding vector:

$$e_i = f_\theta(s_i) \tag{1}$$

Where $\boldsymbol{e}_i$ is the embedding vector for location $i$ and $f_\theta$ denotes the sentence-transformer encoder. These embeddings were stored in a vector database and later used to retrieve historically similar intersections for each test location. The embedding-based retrieval component captures latent semantic similarity between intersections by jointly encoding crash narratives and structured crash descriptors into a shared vector space. As a result, locations exhibiting similar crash mechanisms may be retrieved even when they differ in specific wording.

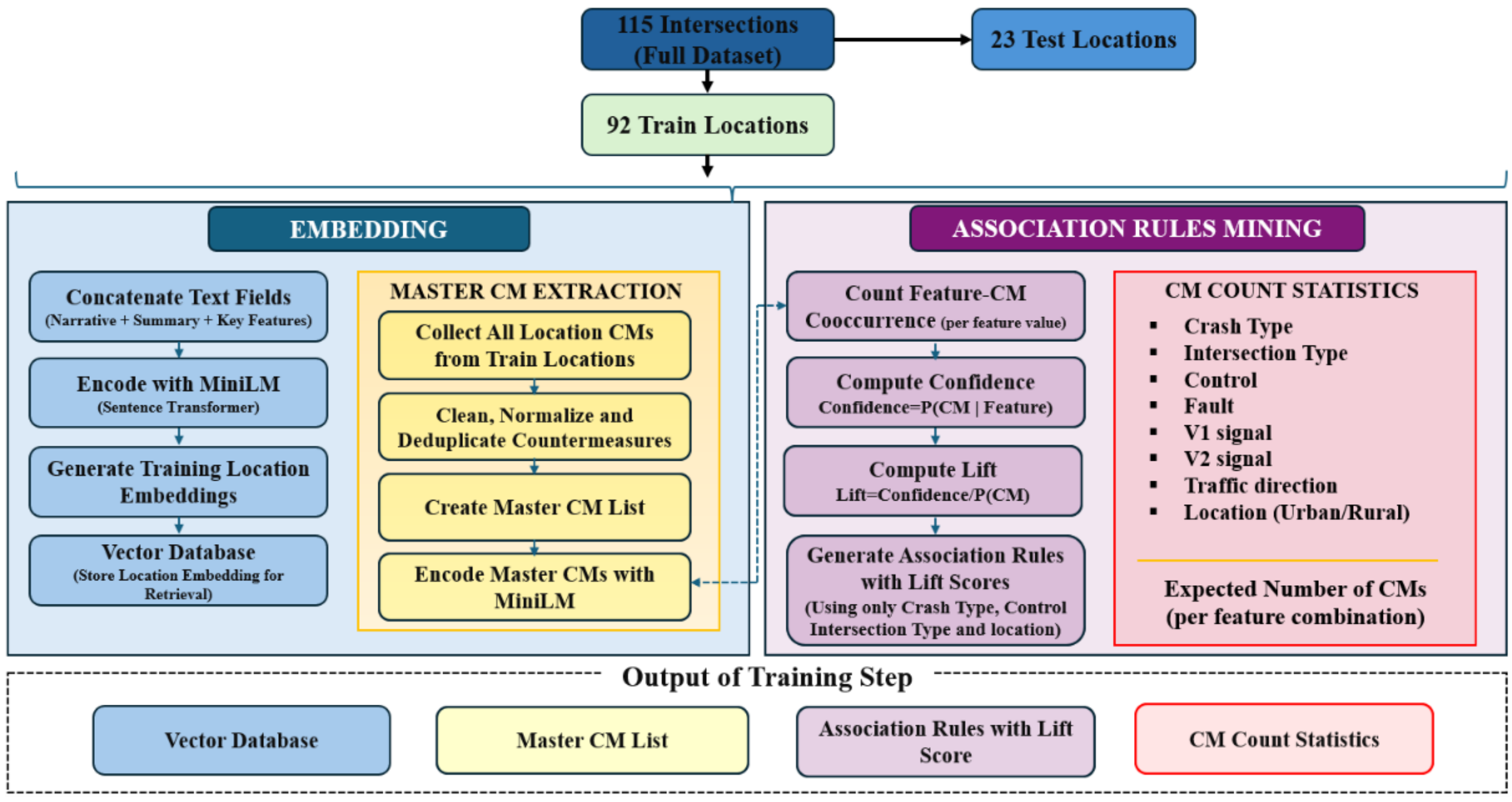

**Figure 4. Training data preparation process**

In parallel, all location-level countermeasures from the training locations were collected, cleaned, normalized, and deduplicated to create a master countermeasure list. This list defined the allowable recommendation space and reduced inconsistencies in wording across locations. Each master countermeasure was also encoded using the same embedding model, allowing generated countermeasure text to be mapped back to the closest standardized countermeasure during post-processing.

Association rule mining was then applied to capture empirical relationships between crash features and countermeasures. For a crash feature (f) and countermeasure (c), confidence was computed as:

$$\text{Confidence}(f \rightarrow c) = \frac{N(f,c)}{N(f)} \tag{2}$$

Where $N(f,c)$ is the number of records in which feature $f$ and countermeasure $c$ co-occurred, and N(f) is the total number of records containing feature $f$. Lift was computed as:

$$\text{Lift}(f \rightarrow c) = \frac{\text{Confidence}(f \rightarrow c)}{P(c)} \tag{3}$$

where $P(c)$ is the overall probability of countermeasure $c$ in the training data. A higher value of lift indicates that the countermeasure is more strongly associated with the feature than expected by chance. Unlike the embedding-based retrieval component, which identifies globally similar intersections using semantic similarity, association rule mining captures localized relationships between crash characteristics and countermeasures. Consequently, the retrieved locations and association rules provide complementary sources of evidence during the recommendation process. In addition, countermeasure count statistics were calculated for combinations of crash type, intersection type, traffic control, vehicle fault, signal status, traffic direction, and urban or rural context. These statistics provided guidance on the expected number of countermeasures for a given crash context. Overall, from this phase produced outputs formed the knowledge base used in the subsequent retrieval-augmented countermeasure recommendation stage.

### 3.4 Retrieval-Augmented Generation (RAG)

The RAG process consists of three core stages: retrieve, augment, and generate. In the retrieval stage, the input query is used to search an external knowledge base and identify relevant supporting information. In the augmentation stage, the retrieved evidence is combined with the original query and other structured knowledge sources to construct an enriched prompt. In the generation stage, the augmented prompt is provided to a large language model (LLM), which produces an output grounded in the retrieved information rather than relying only on the model's parametric knowledge. This process is especially important for transportation safety applications, where countermeasure recommendations should be traceable to observed crash patterns, similar historical locations, and predefined countermeasure options rather than being generated freely by the LLM. **Figure** illustrates the overall pipeline of RAG.

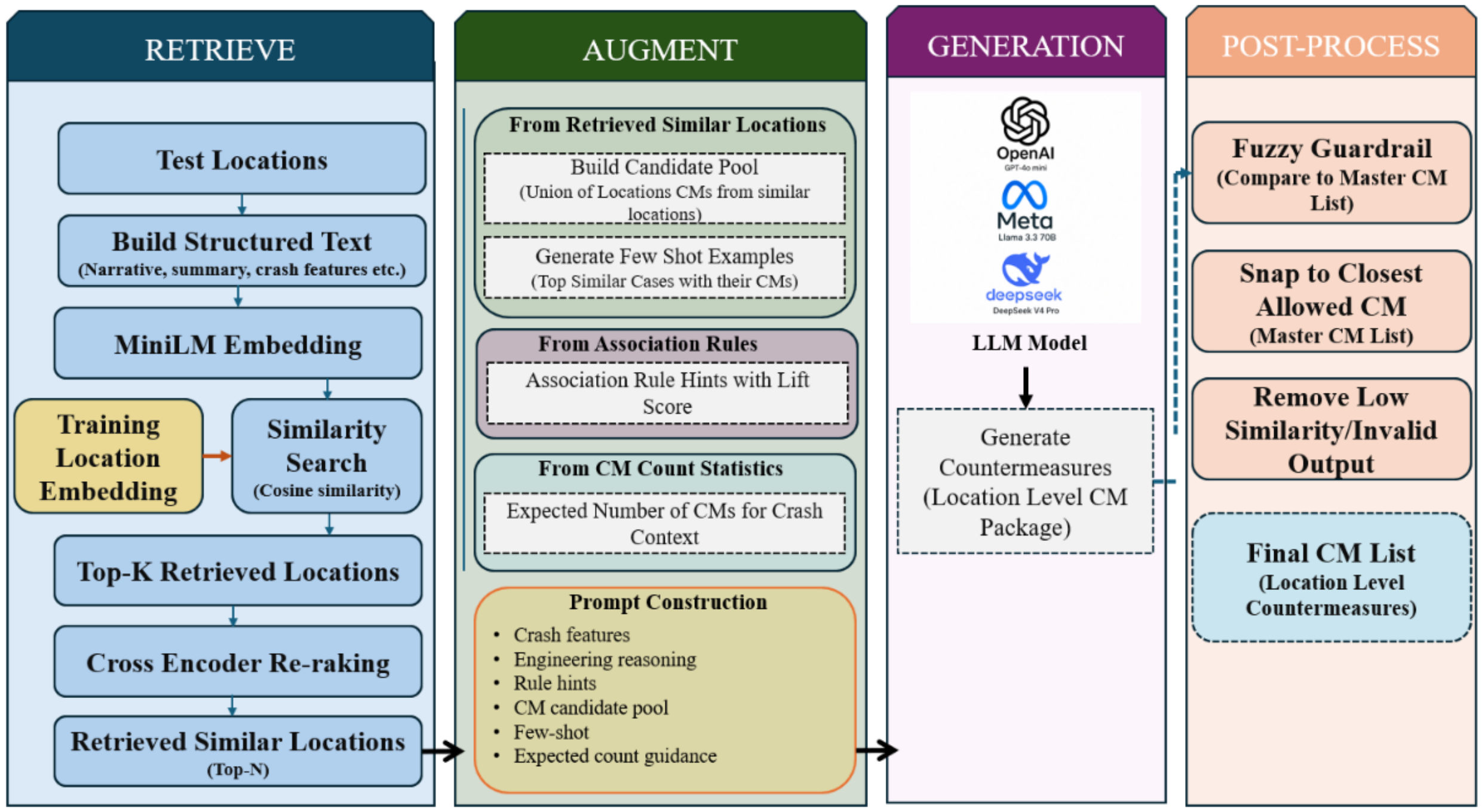

**Figure 5. RAG workflow for countermeasure recommendation**

The objective of the retrieve stage was to identify historically similar locations from the training dataset for each test location. For a given test location, the same structured textual representation used during training was constructed by combining its crash narratives, summaries, and structured descriptors, allowing both unstructured and structured crash information to contribute to the similarity search.

Let $s_q$ denote the structured text representation of a test location and test location encoded as

$$e_q = f_\theta(s_q) \tag{4}$$

The similarity between the test location and each training location was then measured using cosine similarity (Huang 2008):

$$sim(q,i) = cos(\theta) = \frac{e_q \cdot e_i}{\|e_q\|\|e_i\|} \tag{5}$$

Locations with higher cosine similarity were considered more similar in terms of crash context, movement patterns, contributing factors, and narrative descriptions. To improve retrieval quality, the framework first retrieved a larger set of candidate locations using the MiniLM embedding similarity search. These initially retrieved cases were then re-ranked using a cross-encoder model, cross-encoder/ms-marco-MiniLM-L-6-v2. Unlike the initial embedding model, which independently encodes the test and training locations, the cross-encoder jointly evaluates each test candidate pair and produces a more precise relevance score. The final retrieval score combined the original embedding similarity and the cross-encoder score:

$$S_i = 0.3\, sim\,(q,i) + 0.7\left(\frac{r_i}{10}\right) \tag{6}$$

where $S_i$ is the final combined retrieval score for candidate location $i$, and $r_i$ is the cross-encoder re-ranking score. The top-ranked locations after re-ranking were retained as the retrieved similar historical locations. These locations provided empirical evidence for the subsequent augmentation step, including their location-level countermeasure packages and crash characteristics.

After retrieving the most similar historical locations, the retrieved information was transformed into several forms of evidence to support countermeasure generation. The augmentation stage was designed to provide the LLM with a constrained and evidence-based context rather than allowing it to generate countermeasures solely from the crash narrative. Three main sources of augmentation were used: retrieved similar locations, association-rule hints, and countermeasure count statistics. First, the retrieved similar locations were used to construct a candidate countermeasure pool. Each retrieved location contained a compiled location-level countermeasure package, denoted as LocationCMs. The countermeasures from the retrieved locations were extracted, deduplicated, and mapped to the closest valid entry in the master countermeasure list. Let $R_q$ denote the set of retrieved locations for test location $q$, and let $C_i$ denote the set of location-level countermeasures associated with retrieved training location $i$. The candidate countermeasure pool for location $q$ can be expressed as:

$$A_q = \bigcup_{i \in R_q} C_i \qquad (7)$$

Where $A_q$ represents the allowable candidate countermeasure set derived from similar historical locations. Each candidate countermeasure was further checked against the master countermeasure list using semantic similarity. Countermeasures sufficiently similar to a master-list entry were retained and replaced with the standardized master-list wording. This step reduced wording variation and ensured consistency in the final recommendations. In addition to forming the candidate pool, the retrieved locations were also used to generate few-shot examples. These examples provided the LLM with concrete historical precedents showing how similar crash configurations were linked to countermeasure packages.
The second augmentation source was the association-rule mining output from the training step. For each test location, the corresponding crash types, controls, intersection types and locations were used to retrieve relevant rule hints. These hints consisted of countermeasures with their corresponding lift scores. The resulting rule hints can be represented as:

$$H_q = \{(c, L_c): c \in C, L_c > 0\} \qquad (8)$$

Where $H_q$ is the set of rule-based countermeasure hints for location $q$, $c$ is a candidate countermeasure, and $L_c$ is its lift score. These hints were not treated as mandatory recommendations. Instead, they were provided to the LLM as statistical evidence indicating which countermeasures were historically associated with similar crash features. The third augmentation source was the countermeasure count statistics. This component estimated the expected number of countermeasures for a given crash context. The count statistics were learned from the training data using crash-level countermeasures and feature combinations. For each feature combination, empirical statistics such as the mean, minimum, maximum, and percentile-based countermeasure counts were computed. During inference, the most specific available feature combination was used first, and the procedure gradually backed off to less specific combinations when sample size was insufficient. The purpose of the count statistics was not to determine which countermeasures should be selected, but to guide the approximate size of the recommended countermeasure package. This helped prevent the LLM from generating either too few recommendations or an unrealistically large list of countermeasures. The final count guidance provided to the LLM specified an expected range of countermeasures based on historical cases with similar crash contexts. Finally, the retrieved examples, candidate countermeasure pool, association-rule hints, countermeasure count guidance, structured crash features, and crash narrative were combined into a single augmented prompt. The augmented prompt also incorporated an engineering-guided reasoning framework that represented the domain decision process used by safety analysts. Rather than selecting countermeasures directly from statistical hints, the LLM was instructed to first follow a hierarchical logic, starting from crash type, then considering intersection type, control, conflict direction, driver fault, and urban/rural context before evaluating candidate countermeasures. This allowed the model to use retrieved cases and association-rule evidence within a human-interpretable engineering decision structure.

In the generate stage, the augmented prompt was provided to the LLM to produce the location-level countermeasure recommendation. The generation task was formulated as a constrained selection problem. Given a test location $q$, the LLM received the structured crash features, the crash narratives, retrieved historical examples, the candidate countermeasure set $A_q$, rule-based hints $H_q$, and the expected countermeasure count range. The model was instructed to identify the primary crash mechanism first and then select only countermeasures that were consistent with that mechanism. This was designed to prevent recommendations that were statistically associated with a feature but did not logically address the crash mechanism. The generation logic followed three main decision gates. First, the mechanism filter required the model to determine whether each candidate countermeasure directly addressed the identified crash mechanism. Countermeasures that did not address the mechanism were rejected even if they had high lift scores. Second, among the mechanism-consistent candidates, association-rule hints were used to prioritize countermeasures with stronger historical evidence. Third, retrieved similar locations were used to fill remaining gaps by identifying countermeasures that appeared in comparable historical cases. The count guidance was used to control the approximate size of the final recommendation package. The raw output from the LLM was a semicolon-separated list of recommended countermeasures. This list represented the initial generated location-level countermeasure package before post-processing.

Because LLM-generated text may contain minor wording variations or unsupported recommendations, a post-processing stage was applied to ensure consistency with the predefined countermeasure database. The post-processing step compared each generated countermeasure against the candidate countermeasure pool and the master countermeasure list using semantic similarity. This step functioned as a fuzzy guardrail that constrained the final output to valid countermeasure wording.

Let $\hat{c}_j$ denote the *j-th* countermeasure generated by the LLM and let $A_q$ denote the allowable countermeasure pool for test location $q$. Each generated countermeasure was encoded and compared with the allowable candidate countermeasures using cosine similarity. The closest allowable countermeasure was identified as:

$$c_j^* = \arg\max_{c \in A_q} \frac{v(\hat{c}_j) \cdot v(c)}{\|v(\hat{c}_j)\| \, \|v(c)\|} \tag{9}$$

where $v(\cdot)$ represents the embedding vector of a countermeasure text. The generated countermeasure was retained only if its highest similarity score exceeded a predefined guardrail threshold τ (0.80):

$$\max_{c \in A_q} [\text{sim}(\hat{c}_j, c)] \geq \tau \tag{10}$$

If this condition was satisfied, the generated wording was replaced with the closest standardized countermeasure from the allowable set. Otherwise, the generated countermeasure was removed. This ensured that the final recommendation did not include hallucinated, unsupported, or low-similarity outputs.

### 3.5 Evaluation

The framework was evaluated at the location level by comparing the predicted countermeasure package with the observed location-level countermeasure package. Precision, recall, and F1-score were computed for each test location. For each test location *q,* if $Pq = p_1, p_2, \ldots, p_m$ is the set of countermeasures predicted and $Gq = g_1, g_2, \ldots, g_n$ denote the observed ground-truth countermeasure package then precision, recall, and F1-score are calculated as:

$$Precision_q = \frac{1}{|P_q|} \sum_{i=1}^{m} match(p_i);\ match\,(p_i) = \begin{cases} 1, & if \max_{g_j \in G_q} sim(p_i, g_j) \geq \tau \\ 0, & otherwise \end{cases} \tag{11}$$

$$Recall_q = \frac{1}{|G_q|}\sum_{j=1}^{n} recovered;\ recoverd\ (g_j) = \begin{cases} 1,\ if \max_{p_i \in P_q} sim(g_j, p_i) \geq \tau \\ 0,\ otherwise \end{cases} \tag{12}$$

$$F_1 = \frac{2 \times Precision \times Recall}{Precision + Recall} \tag{13}$$

Because the task involved recommending a package of countermeasures rather than predicting a single label, additional location-level match indicators were also reported. Specifically, the number and percentage of test locations with at least one, two, three, four, and five matched countermeasures were calculated. The average number of matched countermeasures per location was also reported, along with the average number of actual and predicted countermeasures. Locations with insufficient training support were identified separately and excluded from the prediction-based evaluation.

## 4 Results and Discussion

### 4.1 Framework Performance

The performance of different alternative approaches are compared and summarized in **Table 1** The baseline methods include a classification-based model (Random Forest multi-label), a rule-based recommendation approach (without LLM), a hybrid approach (retrieval pool re-ranked using the classifier and rule-based method), and an ensemble approach based on weighted voting over the four methods. For all baseline methods, the number of recommended countermeasures was capped at five, consistent with the average package size observed across locations and sufficient to capture the majority of relevant treatments without inflating recall through over-recommendation. All results were obtained using five-fold cross-validation to ensure stable performance estimates across the limited number of evaluation locations.

**Table 1. Comparison of countermeasure recommendation methods**

| Method | Precision | Recall | F1-Score | Avg number of matches | Avg Actual CMs | Avg Predicted CMs |
|---|---|---|---|---|---|---|
| Classification | 0.45 | 0.59 | 0.49 | 2.35 | 3.86 | 5 |
| Rules Only | 0.45 | 0.64 | 0.50 | 2.35 | 3.86 | 5 |
| Hybrid | 0.55 | 0.72 | 0.61 | 2.79 | 3.86 | 5 |
| RAG | **0.82** | **0.85** | **0.82** | **3.14** | 3.86 | 3.91 |
| Ensemble | 0.61 | 0.80 | 0.67 | 3.09 | 3.86 | 5 |

The RAG framework achieved the best overall balance between precision and recall, resulting in the highest F1-score of 0.82. It also produced an average of 3.14 correctly matched countermeasures per location while recommending an average of 3.91 countermeasures, which closely matched the average number of actual countermeasures (3.86).

**Figure** compares the number of test locations achieving at least N correctly matched countermeasures across the five evaluated approaches. The RAG framework successfully identified at least one correct countermeasure for all 22 locations and achieved two or more correct matches for 21 locations. It also identified three or more correct countermeasures for 15 of the 17 eligible locations, four or more correct countermeasures for 7 of the 10 eligible locations, and at least five correct countermeasures for 3 of the 9 eligible locations. The classification model exhibited the weakest performance, particularly for locations requiring multiple countermeasures; although it achieved at least one correct recommendation for 21 locations, the number of locations with three or more correct matches dropped sharply to seven. Rule-based recommendations provided comparable coverage at lower match levels but generated considerably larger recommendation sets, consistent with the lower precision reported in **Table 1**. The hybrid and ensemble approaches further increased the number of locations with multiple matched countermeasures, with the ensemble approach achieving the coverage for 13 locations having three or more correctly matched countermeasures. However, this improvement was accompanied by reduced precision.

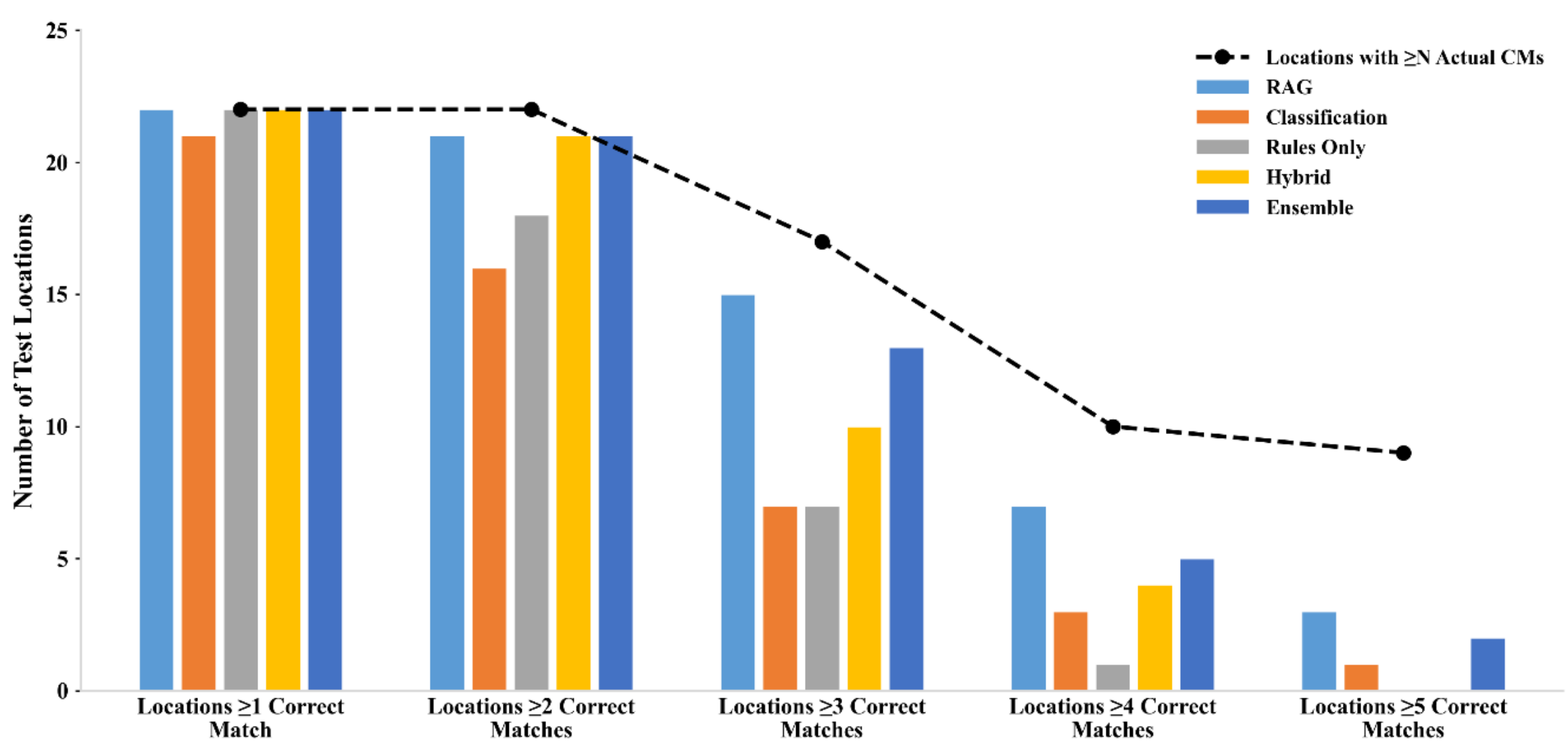

**Figure 6. Comparison of countermeasure matching performance across test locations**

The performance of the RAG framework paired with three different large language models including GPT-4o mini, Llama 3.3 70B, and DeepSeek V4 Pro are compared and summarized in **Figure** . GPT-4o mini achieved the strongest overall performance across nearly all evaluation metrics. It produced the highest precision (0.82), recall (0.85), and F1-score (0.82), indicating a better balance between recommending relevant countermeasures and recovering the observed countermeasure set. Llama 3.3 70B ranked second overall, with an F1-score of 0.75. Although its precision and recall were lower than GPT-4o mini, it identified three or more correct countermeasures at 15 locations, slightly exceeding GPT-4o mini in this single metric. DeepSeek V4 Pro achieved performance comparable to Llama 3.3 70B in terms of precision but exhibited lower recall and F1-score. While all three models successfully identified at least one correct countermeasure for every evaluated location, DeepSeek generated fewer correct recommendations on average and recovered fewer observed countermeasures than the other two models.

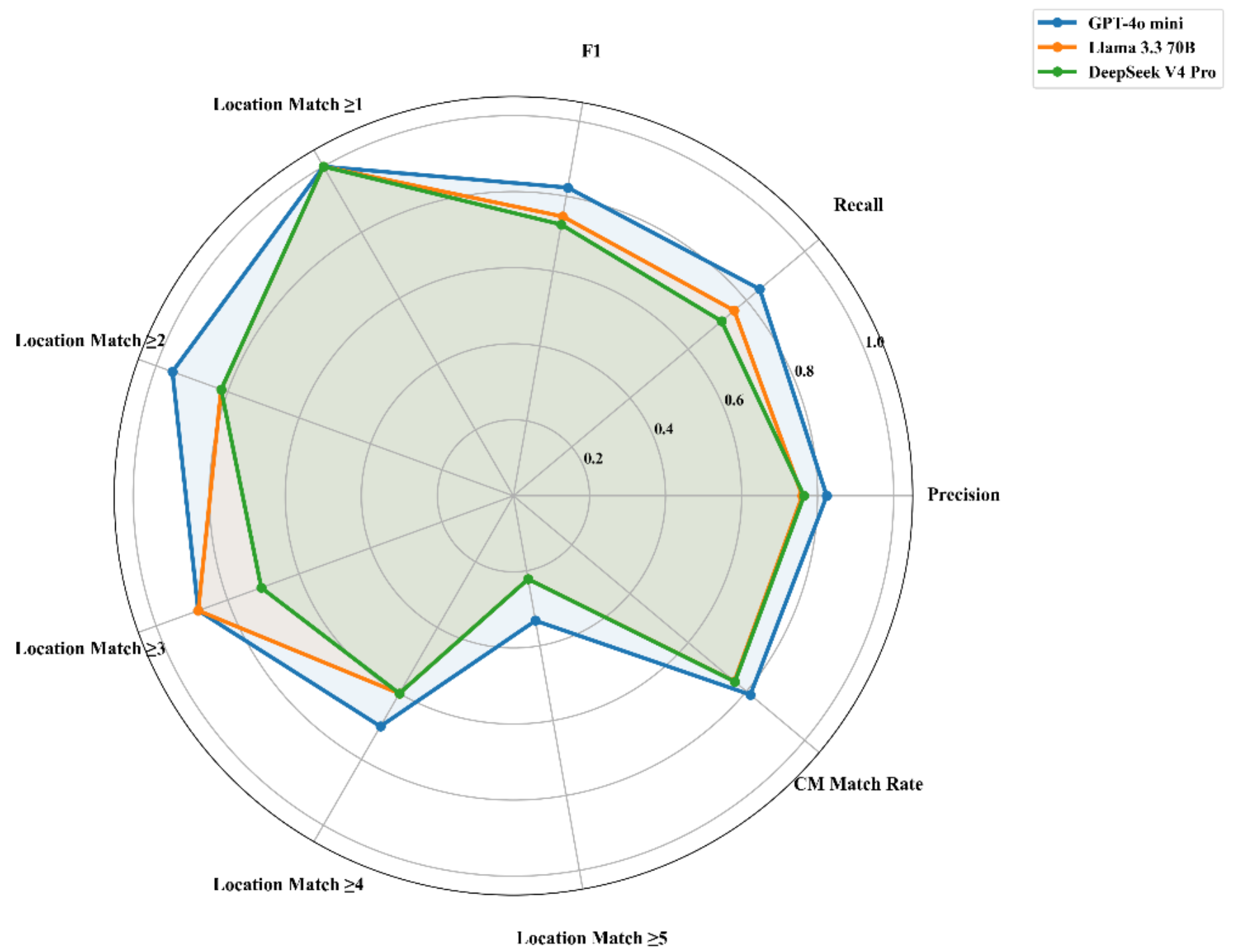

**Figure 7. Performance comparison of different large language models within the RAG framework**

An example illustrating the crash pattern summary of multiple crash narratives and the corresponding LLM-generated countermeasures recommended at an intersection is shown in **Figure 2**.

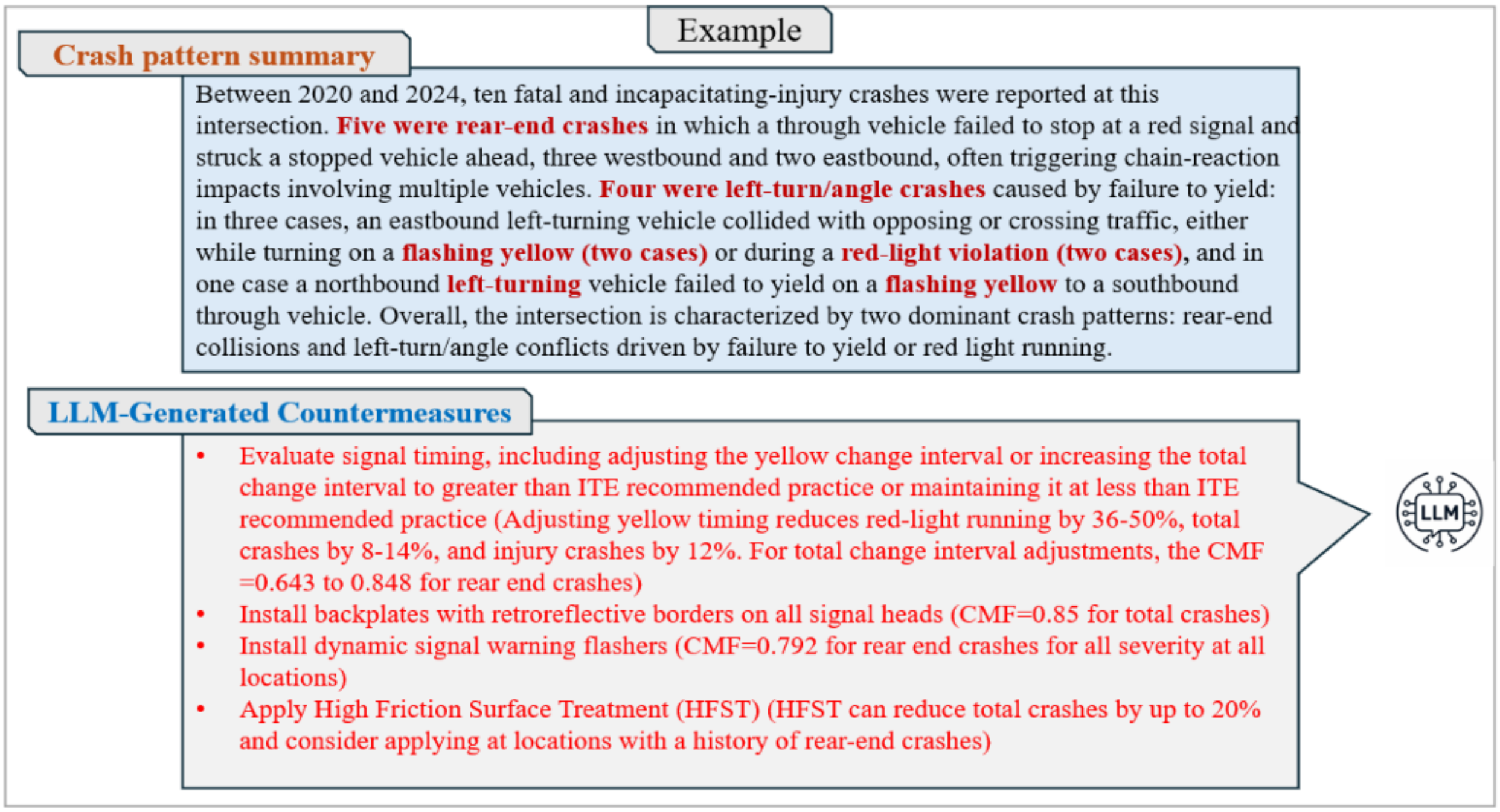

**Figure 2. Example of LLM-generated countermeasures based on crash pattern at an intersection.**

### 4.2 Ablation Study

To assess the contribution of each component within the proposed framework, an ablation study was conducted in which individual components were disabled one at a time while holding all others constant. Performance was evaluated using the same fuzzy-matching metrics applied to the full system. The results are summarized in **Table 2**.

**Table 2. Component wise ablation results evaluating the contribution of each module to overall performance**

| Configuration | Precision | Recall | F1-Score | Avg number of matches | Avg Actual CMs | Avg Predicted CMs |
|---|---|---|---|---|---|---|
| Full System | 0.82 | 0.85 | 0.82 | 3.14 | 3.86 | 3.91 |
| (-) Reasoning framework | 0.80 | 0.82 | 0.80 | 3.09 | 3.86 | 3.86 |
| (-) Three-gate filter | 0.74 | 0.78 | 0.71 | 2.64 | 3.86 | 3.55 |
| (-) Association rules | 0.67 | 0.66 | 0.65 | 2.64 | 3.86 | 3.82 |
| (-) CM count guidance | 0.63 | 0.93 | 0.74 | 3.50 | 3.86 | 5.45 |
| (-) Reranking | 0.74 | 0.75 | 0.73 | 2.82 | 3.86 | 3.86 |
| (-) Few-shot examples | 0.79 | 0.72 | 0.74 | 2.64 | 3.86 | 3.36 |

Removing the association rules produced the largest degradation in performance, with F1-score falling from 0.82 to 0.65 and both precision and recall declining in tandem (to 0.67 and 0.66, respectively). This confirms that the statistically mined associations provide substantial guidance toward appropriate countermeasures, and that their removal forces the language model to rely on retrieval and reasoning alone, which proves insufficient. The three-gate selection mechanism was the second most influential component. Its removal reduced the F1-score to 0.71, with recall dropping from 0.85 to 0.78 and the average number of matched countermeasures falling from 3.14 to 2.64. This demonstrates that the mechanism-first filtering and ranking process meaningfully improves the relevance of selected countermeasures. Disabling the CM count guidance revealed a clear precision–recall trade-off. Without it, the system over-predicted substantially, increasing the average number of recommended countermeasures from 3.91 to 5.45. While this inflated recall to 0.93, since recommending more countermeasures increases the likelihood of matching the ground truth and precision dropped sharply to 0.63. This confirms that the count guidance plays an important role in constraining recommendations to a realistic and focused set. The reranking and few-shot example components contributed more modest improvements, with their removal reducing F1-score to 0.73 and 0.74, respectively. Both primarily affected recall and the average number of matched countermeasures, indicating that they help surface relevant historical precedents that the model might otherwise overlook. Interestingly, removing the engineering reasoning framework resulted in only a marginal change in performance. This suggests that, for the present dataset, the explicit reasoning structure offers limited gains in predictive accuracy. Nonetheless, the framework retains practical value beyond predictive accuracy by enforcing a transparent, domain-consistent reasoning process that makes the system's recommendations more interpretable for practitioners.

## 5 Conclusion

This study presented a crash narrative-guided retrieval-augmented generation framework for recommending site-specific countermeasures at intersections. By extracting crash mechanisms from unstructured narratives including traffic control, signal indication, driver fault, vehicle movements, and travel directions and linking them to evidence-based treatments from the FHWA Proven Safety Countermeasures and the CMF Clearinghouse, the framework translates narrative-derived crash patterns into actionable engineering recommendations. The approach integrates embedding-based retrieval of historically similar intersections, association-rule mining, and an engineering-guided reasoning framework that directs the language model through a domain-consistent decision process before selecting countermeasures. Based on 115 intersections in Lake and Sumter Counties, Florida, and evaluated using five-fold cross-validation, the proposed framework achieved a precision of 0.82, recall of 0.85, and F1-score of 0.82, with an average of 3.14 matched countermeasures per location, closely aligned with the average actual (3.86) and predicted (3.91) countermeasure counts. This performance substantially outperformed classification-based, rule-based, hybrid, and ensemble techniques. The ablation study confirmed that the association rules, three-gate mechanism filter, and countermeasure count guidance were the most influential components, while the reasoning framework contributed interpretability and domain consistency beyond predictive accuracy.

The findings demonstrate that crash narratives, combined with a retrieval-augmented large language model framework, can support transportation agencies in selecting intersection countermeasures more systematically and at scale. Beyond recommendation, the framework highlights the potential of large language models to extract crash mechanism information from unstructured narratives, transforming descriptive crash reports into structured engineering knowledge that, when integrated with established countermeasure resources and engineering reasoning, yields recommendations that are both context-specific and consistent with accepted safety practices. Nevertheless, several limitations remain. The evaluation was based on a limited number of intersections within two counties, and performance may vary across regions with different roadway characteristics and reporting practices. Future work should expand the dataset across diverse jurisdictions, incorporate additional contextual factors such as traffic volume and roadway geometry, and further evaluate the framework as a practical decision-support tool for transportation safety engineers.

## Declaration of competing interest

The authors declare no conflict of interest. All authors reviewed the results and approved the final version of the manuscript.